\documentclass{article}

\usepackage{arxiv}
\usepackage[utf8]{inputenc}
\usepackage[T1]{fontenc}
\usepackage{hyperref}
\usepackage{url}
\usepackage{amsmath}
\usepackage{amsfonts}
\usepackage{booktabs}
\usepackage{graphicx}
\usepackage{biblatex}
\usepackage{cleveref}

\title{Exploring spectro-temporal features in end-to-end convolutional neural networks}

\author{
  Sean Robertson$^{1,2}$, Gerald Penn$^1$, Yingxue Wang$^1$ \\
  $^1$Department of Computer Science, University of Toronto, Canada \\
  $^2$Vector Institute, Canada \\
  \texttt{\{sdrobert,gpenn,yingxue\}@cs.toronto.edu}
}

\begin{document}

\maketitle

\begin{abstract}

Triangular, overlapping Mel-scaled filters (``f-banks'') are the current
standard input for acoustic models that exploit their input's time-frequency
geometry, because they provide a psycho-acoustically motivated time-frequency
geometry for a speech signal. F-bank coefficients are provably robust to small
deformations in the scale. In this paper, we explore two ways in which filter
banks can be adjusted for the purposes of speech recognition. First, triangular
filters can be replaced with Gabor filters, a compactly supported filter that
better localizes events in time, or Gammatone filters, a
psychoacoustically-motivated filter. Second, by rearranging the order of
operations in computing filter bank features, features can be integrated over
smaller time scales while simultaneously providing better frequency resolution.
We make all feature implementations available online through open-source
repositories. Initial experimentation with a modern end-to-end CNN phone
recognizer yielded no significant improvements to phone error rate due to
either modification. The result, and its ramifications with respect to
learned filter banks, is discussed.

\end{abstract}

\section{Introduction}

Time-frequency analyses of speech have long been the dominant feature
representation for speech recognition. There have been many different
transformations that attempt to localize an ``event'' in time and frequency,
such as wavelets and Wigner-Ville distributions. A powerful and persistent form
of analysis is the Short-Time Fourier Transform (STFT). It and its derivatives
treat a speech signal as a series of windowed stationary processes. Even though
neural architectures have proven capable of processing raw speech signals
\cite{Ghahremani2016Acoustic,Sainath2015Learning,Oord2016Wavenet}, they tend to
require far more data, aggressive averaging across learned filters, and even
then still benefit from some filter-based initialization. The best results on
benchmark datasets, such as TIMIT, are still achieved with triangular,
overlapping Mel-scaled log filters -- ``f-banks''
\cite{Toth2014Combining,Ravanelli2017Improving}. F-banks post-process the STFT
by isolating and integrating over different bands of the power spectrum. The
stationary assumption of the STFT combined with cleverly spaced band-pass
filters makes for a very robust time-frequency representation of the speech
signal.

There remains the possibility that, at least in the case of neural
architectures
that do use the time-frequency geometry of the f-bank, what is needed is not
necessarily a different representation, but an improvement upon this already
time-tested representation, such as by improving its time-frequency resolution.
As neural architectures become more sophisticated, a higher resolution may in
fact become more desirable. We can incrementally improve the f-bank formula by
optimizing the time-frequency trade-off while respecting the conventions of
time-frequency features - namely, an audio signal produces a sequence of
``frames'' uniformly sampled in time.

There are many alternatives to the triangular filter that may have desirable
properties for ASR. Two such filters are the Gabor and Gammatone. The former
has a provably optimal time-bandwidth product \cite{Mallat2008Wavelet}, whereas
the latter more closely resembles the stimulus response of the human auditory
system \cite{patterson1987efficient}. Both are much more efficient in terms
of time and frequency resolution, but because because the STFT
pipeline integrates over a uniform window in time, much of that resolution is
sacrificed.%
\footnote{The Uncertainty Principle, as applied to digital
signal processing, states that one cannot have an arbitrarily fine
resolution in both time and frequency
\cite{Mallat2008Wavelet}. Filters with short temporal support will have wide
bandwidths, and very narrowband filters will have longer temporal support.}
The information lost could be
valuable, especially to acoustic models that treat the time-frequency
representation as a 2-D geometry, such as those based upon Convolutional
Neural Networks (CNNs).

A second potential improvement is to swap the order of integration in the STFT
pipeline. This new order would more faithfully represent both the filters' time
and frequency bandwidths, and can integrate over shorter time scales, since it
would avoid incurring a resolution penalty by directly windowing. Together,
these two improvements would make a filter bank with better time and frequency
resolution, though they could also be employed separately.

This paper's contributions are twofold: first, we present the aforementioned
modifications to speech features in detail. These modifications, alongside
more traditional speech features, are made available in the accompanying
open-source Python package%
\footnote{\url{https://github.com/sdrobert/pydrobert-speech}}%
as well as through the open-source Matlab repository COVAREP
\cite{g.degottexCOVAREPCollaborativeVoice2014}. Second, we separately evaluate
these adaptations in the framework of an end-to-end speech recognition task.
Constructing a CNN from the architecture described in
\cite{Zhang2016Towards} for the TIMIT phone recognition task, we decode the
phone sequence internally using Connectionist Temporal Classification (CTC)
\cite{Graves2006Connectionist}. We discuss the position of our results as it
relates to current trend of learning filter banks from the raw waveform. The
code developed for this experiment, including the trained weights, are also
open-source and available online
\footnote{\url{https://github.com/sdrobert/more-or-let}}.

\section{Mel-scaled log filter banks} \label{sec:mel}

To illustrate what a coefficient of an f-bank captures in time and frequency, we
adapt an argument from \cite{Anden2014Deep}. In the continuous domain, a filter
bank coefficient $k$ for a given frame of length $T$ centered at sample $c$, is
popularly calculated for signal $f$ as \cite{Young2006Htk,Povey2011Kaldi}:
\begin{equation} \label{eq:fbank}
\mathtt{FBANK}_{T,f,w}(c, k) = \log
  \int_{-\infty}^{\infty} \widehat{h_k}(\omega)
  \left|\widehat{fw_{c,T}}\right|^2(\omega) d\omega
\end{equation}
Where $\hat{x}$ is the Continuous Fourier Transform (CFT) of $x$, $h_k$ is the
$k$\textsuperscript{th} real filter in the filter bank, $w_{c,T}$ is a windowing
function of temporal support $T$ centered at $c$.

The $h_k$ are triangular windows with vertices derived by linearly
sampling the Mel scale \cite{Stevens1937Scale}. The Mel scale, based
on psychoacoustic experimentation, is roughly linear with respect to
frequency below 1000Hz and logarithmic above. Clearly, the frequencies
captured by a coefficient of the f-bank are limited to the nonzero
region of $\widehat{h_k}$. Because the bandwidth of $h_k$ scales with
its central frequency, f-banks are robust to time warping
\cite{Anden2014Deep}. In contrast, acoustic models that operate
directly on the normalized power spectrum
\cite{Collobert2016Wav2Letter,Amodei2016Deep} are sensitive to time-warping.
Without log-scale spacing, power spectrum models must repeatedly learn the
same patterns, transformed by small-scale dilations ($T_\alpha f(i) = f(\alpha
i)$) that are commonly introduced by differences in vocal tract length
\cite{Irino2002Segregating}.

Multiplying the signal with a window in time blurs the filtered frequency
responses. By the convolution theorem, $\widehat{fw_{c,T}} = \widehat{f} \ast
\widehat{w_{c,T}}$. For large $T$, $\widehat{w}(\omega) \rightarrow
\delta(\omega)$, the Dirac delta function ($\int_{0^-}^{0^+} \delta(\omega)
d\omega = 1$ and $\delta(\omega) = 0$ for $\omega \neq 0$). Nonetheless,
$\widehat{w}$ will have a nonzero support. This functions similarly to a
rolling average over $\widehat{f}$ and effectively increases the bandwidth of
all $h_k$. While not particularly detrimental to wideband filters, the bandwidth
of the window can be similar or greater than that of the narrowband filters,
which can drastically reduce their discriminative power.

The $h_k$ are also entirely real in the Fourier domain. We can instead let
$\sqrt{h_k}$ represent a filter with a frequency response equal to the
point-wise square root of $h_k$'s, and shift the filter into the modulus. By the
convolution theorem and Parseval's theorem, \Cref{eq:fbank} can be rearranged
into an integration over the filtered response of the window:
\begin{equation} \label{eq:fbank_time}
\mathtt{FBANK}_{T,f,w}(c, k) = C + \log\int_{-\infty}^\infty \left|\sqrt{h_k}
\ast (fw_{c,T}) \right|^2(t) dt
\end{equation}
If $w_{c,T}$ is distributed non-uniformly in time, even local transformations of
$f$ will affect the resulting coefficient. Popular choices of $w_{c,T}$, such as
Hann, Hamming, or triangular windows, attenuate values at the periphery of the
window. In effect, signal transients smoothly transition between overlapping
frames. However, a narrowband $\sqrt{h_k}$ decays more slowly to zero
in time and has a number of cusps in its envelope (side lobes). This makes the
integration difficult to interpret temporally.

\section{Filter types} \label{sec:filters}

Triangular or square-root filters are not the only types of filters that can be
scaled in the Mel-domain. The first modification to the spectro-temporal
recipe that we explore is to the type of filter employed. We experiment with
two additional types of filter: the Gabor filter and the Gammatone filter.

\subsection{Gabor filters} \label{sec:gabor}

Gabor filters have been explored in a variety of contexts. 2-D Gabor filters
are often applied to spectrograms (or log-spectrograms) for ASR as a collection
of spectral-temporal features
\cite{Schaedler2015Separable,Chang2013SpectroTemporal,Domont2008Hierarchical}.
Their 2-D construction allows the bank to capture meaningful geometric
structures, such as formant movement. Likewise, 2-D Gabor bases have been
learned as convolutional layers \cite{Chang2014Robust}. The present paper
focuses on the design and evaluation of spectrogram-like features, rather than
high-level spectro-temporal features that sit atop a spectrogram. In
\cite{Dimitriadis2011On}, a Mel-scaled Gabor filter bank was designed much like
the one presented here, but it was 1-D and employed in an HMM-based
architecture, not a neural network. Recently,
\cite{zeghidourLearningFilterbanksRaw2018} trained end-to-end CNNs for
phone recognition with weights initialized to a Gabor filter bank.

Gabor filters are simply Gaussian windows with a complex carrier. They
are defined in time as
\begin{equation} \label{eq:gabor_time}
h_k(t) = C e^{-\frac{t^2}{2s_k^2} + i\xi_kt}
\end{equation}
and frequency as
\begin{equation} \label{eq:gabor_freq}
\widehat{h_k}(\omega) = \sqrt{2}\pi\sigma^{3/2} C
                        e^{\frac{-\sigma^2(\xi_k - \omega)^2}{2}}
\end{equation}
To design the filter bank, center frequencies $\xi$ are sampled along the
Mel-scale. Neighbouring filters' frequency supports intersect at their -3dB
bandwidths. Gabor banks (g-banks) are calculated the same way as in
\Cref{eq:fbank_time}, excluding the point-wise square root.

Gabor filters have a provably optimal time-frequency trade-off
\cite{Mallat2008Wavelet}. Their regions of effective support in both time and
frequency are bounded above by a Gaussian window.

One downside to the Gabor filter is its symmetric frequency response.
Log-linear scales of speech perception, such as the Mel scale, are decidedly
asymmetric. The Gammatone filter helps mitigate this trade-off.

\subsection{Gammatone} \label{sec:gammatone}

Gammatone filters were derived in
\cite{aertsenSpectrotemporalReceptiveFields1981}. Their skewed frequency
response lend themselves nicely to existing models of speech perception
\cite{patterson1987efficient}. Gammatones have been employed in
ASR directly \cite{r.schluterGammatoneFeaturesFeature2007,y.shaoAuditorybasedFeatureRobust2009}
or used as a starting point in learned feature representations
\cite{Zeghidour2018,Sainath2015Learning}.

The complex Gammatone filter is defined in time as
\begin{equation} \label{eq:gamma_time}
  h_k(t) = C t^{n-1} e^{-\alpha t + i \xi_k t} u(t)
\end{equation}
Where $n$ is the \textit{order} of the Gammatone, usually set to 4, which
controls the skewness of the envelope of $h$. The filter is defined in
frequency as
\begin{equation} \label{eq:gamma_freq}
  \widehat{h_k}(\omega) = \frac{C(n - 1)!}
                          {\left( \alpha + i (\omega - \xi_k)\right)^n}
\end{equation}
The same strategy for designing g-banks can be applied to Gammatone filter
banks (tone-banks).

The Gammatone does not have an optimal time-frequency trade-off like the
Gabor filter. It is a first-order approximation of the Gammachirp
filter, which is time-\textit{scale} optimal
\cite{toshioOptimalAuditoryFilter1995}. Incorporating the Gammachirp into
the standard time-frequency pipeline is difficult, however, because the
Gammachirp is not a linear-time invariant system.

\section{Short integration} \label{sec:si}

As is shown in \Cref{sec:mel}, windowing widens the bandwidth of the narrowband
filters in the f-bank, a form of spectral leakage. Increasing the width of the
window will decrease the magnitude of the spectral leakage. However, a wider
window will capture more of the signal in time, decreasing its temporal
resolution. By taking inspiration from scattering transforms
\cite{Mallat2010Recursive,Anden2014Deep}, we can modify f-bank computations to
better capture low-frequency information.

A sample of a first-order 1-D scattering transform is:
\begin{equation} \label{eq:scat_sample}
\mathtt{SCAT}_{T,f,\psi,\phi}((\lambda_1),u) = \frac{1}{T}
  \int_{-\infty}^\infty|f \ast \psi_{\lambda_1}|(t)\phi(c - t)dt
\end{equation}
for some family of wavelets $\psi$ with $\psi_\lambda(t) =
\lambda^{-1}\psi(\lambda^{-1}t)$, and $\phi$ a low-pass filter.
Second-order scattering transforms are all paths of length 2 of the form $||f
\ast \psi_{\lambda_1}| \ast \psi_{\lambda_2}|$, and so on. The key observation
is that the low-pass window has been shifted out of the nonlinear region. Note
that
\begin{equation*}
|f \ast \psi|^2(t) = \left(f \ast \psi\right)\overline{\left(f \ast \psi\right)}(t)
  \mapsto \left(\hat{f}\hat{\psi} \ast \overline{\hat{f}}\overline{\hat{\psi}}\right)
\end{equation*}
For real-valued $f$ and $\psi$, this is an autocorrelation in frequency. In
general, the modulus pushes high-frequency information towards zero, where it
can be captured by the low-pass $\phi$.

Existing work that uses scattering in speech has focused on wide $\phi$ and
high-order (greater than first) paths
\cite{Peddinti2014Deep,Anden2014Deep,Zeghidour2016Deep}. Furthermore, though a
scattering path is a contractive operation with a normed space as its image,
the
modulus is nonlinear. The nonlinearity makes it difficult to patch together a
cohesive 2-D geometry along higher-order paths. Thus, scattering transforms are
most useful in classifying stationary processes \cite{Mallat2016Understanding}.

Instead of using a cascade of filters to capture high-frequency information, we
can simply repeat the process in \Cref{eq:scat_sample}, but with a shorter
window. The following is the Short-Integration Filter Bank (sif-bank):
\begin{equation} \label{eq:si_bank}
\mathtt{SIF\_BANK}_{f,w}(k, c) = \log \int_{-\infty}^{\infty} |f
\ast \sqrt{h_k}|^2w_{c,T}(t)dt
\end{equation}
Windowing still performs a rolling average in the frequency domain, but,
crucially, it is performed after the convolution. All energy within the modulus
originated from the bandwidth of interest (dictated by $\sqrt{h_k}$). Hence,
the sif-bank produces a more accurate representation of the frequency domain.

Since the filter bank need only be invariant to small transformations
\emph{between} frames, we can choose $T$ to be proportional to the frame shift.
More complicated choices of $T$ based on the window type, however, can lead to
improved time resolution without introducing significant aliasing. Treating
$\mathtt{SIF\_BANK}(\cdot, c)$ as a discretely-sampled signal over sampling
interval $i \rightarrow t \Delta c$. Applying this mapping and rearranging
\Cref{eq:si_bank}, we can view a coefficient of the sif-bank as discretely
sampling a continuous distribution:
\begin{equation*}
  \mathtt{SIF\_BANK}_{f,w,k}[i] = \left(\log |f
    \ast \sqrt{h_k}|^2 \ast w_T \right)[i]
\end{equation*}
For a fixed frame shift $\Delta$ seconds, the Nyquist-Shannon Sampling Theorem
dictates that the sif-bank cannot represent frequencies higher than
$1/(2\Delta)$ Hertz unambiguously. Frequencies above said limit will be
subject to aliasing. The size of the window $T$ can be chosen to
(approximately) enforce this limit. One may adjust $T$ until the
zero-crossing or -3dB bandwidth of the main lobe of the window's frequency
response matches $1/(2\Delta)$ Hertz.

Decreasing the size of both the frame shift interval
and window size can increase the temporal resolution of a short-integration
filter coefficient when the bandwidth of the window is much smaller than the
bandwidth of the modulus in \Cref{eq:si_bank}. A ``downside'' of
short-integration is that decreasing the window size will not increase
temporal resolution when the bandwidth of the modulus is smaller than that of
the window already, as is the case for narrowband filters in the bank. For the
low-frequency narrowband filters in a psychoacoustic filter bank, windowing
has little effect, as the temporal support of these filters already far
exceeds that of the frame shift. Thus, the coefficients corresponding to a
narrowband filter in a sif-bank have poorer temporal resolution than those
in an f-bank. This apparent downside is specious: the poor temporal
resolution of narrowband filters is an inescapable property of the
Uncertainty Principle, not the short-integration process. Any improved
temporal resolution in STFT-based filter bank features is paid for with a
less faithful frequency representation, discussed in \Cref{sec:mel}.

It is worth noting that, due to the poor temporal resolution of narrowband
filters, their coefficients will be more highly correlated across time
than wideband filters. This makes the sampling along those bands highly
redundant when the frame shift is small. Redundant representations are not
necessarily a bad thing for classification; sequential frames of f-banks are
overlapping in time, for example. Deltas and double deltas are merely
convolutions of f-bank coefficients, which will clearly correlate the frames.
Previously, f-banks were shown to be more effective than cepstral coefficients
for deep learning precisely because the coefficients of the former are more
strongly correlated \cite{Mohamed2012Understanding}.

\Cref{eq:si_bank} generalizes to arbitrary choices of filter banks. Thus, we
can combine the Gabor filters (\Cref{sec:gabor}) and Gammatone filters
(\Cref{sec:gammatone}) with short-integration to generate Short-Integration
Gabor Banks (sig-banks) and Short-Integration Gammatone Banks (sitone-banks).
The theoretical benefit of Gabor filters is their increased temporal resolution
over f-banks, taking far less time to decay to near zero, whilst Gammatone
filters are more faithful to human auditory perception.

Lastly, we address the computational efficiency of the short-integration
filter bank. STFT-based computations have identical computational
complexity with short-integration $O(N \log N)$. The short-integration
implementations in are
FFT-based, performing the overlap-save method of convolution. Because the
FFT is not preceded by windowing, short-integration requires much larger
FFTs to avoid the effects of circular convolution. This, and the required
inverse FFT, increases computational time. Fortunately, a corresponding
decrease in computational time comes from performing fewer FFTs in total.
Traditional filter banks require an FFT for each frame of coefficients, since
each frame windows the signal anew, whereas short integration can leverage
coefficients from a single FFT-iFFT in multiple frames. Hence, overall
computational times between the types of computation are quite comparable.

\section{Experiment} \label{sec:exp}
In order to explore the efficacy of filter types and methods of computation in
speech recognition, we tested them as drop-in replacements for f-banks in a
deep end-to-end recognizer designed for f-banks. The TIMIT phone recognition
task allows for fast experimental comparison and reduces the impact of language
modeling on experimental results.

\subsection{Data}
We used Kaldi's TIMIT recipe \cite{Povey2011Kaldi} to partition the audio data.
TIMIT's core test set is comprised of 192 utterances of 24 unique speakers. A
50 speaker set of 400 utterances is peeled from TIMIT's complete test set for
early stopping. 462 speakers and 3969 utterances comprise the training set.
``Dialect sentences'' (SA entries) were removed. The full set of 61 phone
labels (including glottal closure) were used for training and decoding, but
collapsed to the standard 39-phone set when calculating Phone Error Rate (PER),
which includes the silence phone.

\subsection{Model} \label{sec:model}
The acoustic model, developed with Keras \cite{Chollet2015Keras} and
Tensorflow \cite{Abadi2015Tensorflow}, mirrors the one described in
\cite{Zhang2016Towards}. The model comprises of mostly convolutional layers
with maxout activations \cite{Goodfellow2013Maxout}. Connectionist Temporal
Classification (CTC) \cite{Graves2006Connectionist} acts as the loss function
for the network. CTC embeds the forward-backward algorithm into backpropagation
to output phone labels directly, curtailing the need for external sequence
modeling (e.g. Hidden Markov Models).

Maxout activations take the per-unit maximum of the output of at least
two weight matrices that have received the same input. This (at least)
doubles the number of trainable weights in memory. After discussion
with the first author of \cite{Zhang2016Towards}, we halved the weights
listed therein to fit the 4.3 million parameter point listed
with a size-2 maxout function.

Following \cite{Zhang2016Towards}, the bottom-most 10 layers of the network
convolve intermediate representations with a $5 \times 3$ (time $\times$
frequency) kernel with stride 1. There is only one pooling layer: a max pooling
of size $1 \times 3$ after the bottom-most convolutional layer. The first four
convolutional layers have 64 feature maps; the remainder have 128. Above the
convolutional layers sit 3 time-distributed fully connected layers. The
frequency axis is collapsed into the feature map axis with a max operator, and
each such vector, indexed by time, is multiplied with the same fully-connected
weight matrix. Each layer has 512 time-distributed hidden units. A final
time-distributed weight matrix constructs the activation matrix over the 61
phone labels for CTC.

\subsection{Training and decoding}
In the first stage of training, optimization is performed with Adam
\cite{Kingma2014Adam} at a learning rate of $10^{-5}$. Dropout
\cite{Srivastava2014Dropout} of probability $0.3$ is applied after the
activation function of each layer. The stage ends when a model's validation loss
has not improved for 50 epochs. Afterwards, optimization continues using
Stochastic Gradient Descent (SGD) at a learning rate of $10^{-8}$ and an L2
weight regularization penalty of $10^{-5}$. The early stopping regime is the
same as in the first stage. Weights are saved after each epoch, and the weight
set with the lowest validation loss from the second stage is used to decode.
Initially, we tuned the beam width used in decoding on the development set.
However, we quickly found that larger beam widths were almost always
preferable, so we fixed the width to 100. Note that development of the CNN was
always performed using f-banks; no architectural or optimization decisions were
influenced by the experimental filters.

\subsection{Features} \label{sec:feats}
The model is trained on four feature sets of identical shape. f-bank is
our implementation of the standard log Mel-scaled triangular filter bank,
g-bank refers to the log Mel-scaled Gabor filter bank proposed in
\Cref{sec:gabor}, tone-bank refers to the log Mel-scaled Gammatone
filter bank from \Cref{sec:gammatone}, and sif-bank, sig-bank, and sitone-bank
are the short-integration analogues (introduced in \Cref{sec:si}) of f-bank,
g-bank, and tone-bank, respectively.

For the standard STFT pipeline features (f-bank, g-bank, and
tone-bank), 40 log filters plus one energy coefficient are calculated
every 10ms over a frame of 25ms. The short-integration filters' window size was
chosen to be 20ms, i.e., double the frame shift, in order to avoid aliasing.
Filters are spaced uniformly on the Mel-scale between 20Hz and 8000Hz. Deltas
and double deltas are concatenated to the end of each frame vector, totalling
123 dimensions. Pre-emphasis, dithering, and compression were enabled at their
standard Kaldi values. An additional baseline, kaldfb, was included to
test Kaldi's built-in f-bank implementation as a sanity check.

\subsection{Evaluation} \label{sec:eval}

To evaluate the performance of the filters and computations, we performed a
hybrid non-parametric statistical analysis. First, we performed a Friedman
test over the four filter types: kaldifb, f-bank, g-bank, and tone-bank.
The Friedman test is approprate for repeated measures of ranked data with more
than two levels when the distribution of the dependent variable (in this case
PER) is not assumed to be Gaussian. For the filter bank with the lowest mean
PER, we performed a Wilcoxon signed-rank test between PERs derived from
regular computations versus those derived from short-integration computation.
10 trials were performed with different seeds for each combination of filter
and computation, for a total of 70 trained models. One seed - the same seed
for each combination of filters and computations - failed to converge, with
PERs around $70\%$. Those trials were removed from analysis, leaving 9 trials
each.

\section{Results and discussion} \label{sec:results}

\begin{table}
\centering
\begin{minipage}{0.47\textwidth}
  \centering
  \begin{tabular}{r l}
    \toprule
    \textbf{Filter} &  \textbf{Test PER (std)} \\
    \midrule
    kaldifb & 18.82 (0.14) \\
    f-bank & 18.60 (0.22) \\
    g-bank & 18.77 (0.26) \\
    tone-bank & 18.71 (0.27) \\
    sif-bank & 18.74 (0.26) \\
    sig-bank & 18.68 (0.36) \\
    sitone-bank & 18.75 (0.30) \\
    \bottomrule
  \end{tabular}
  \caption{
    Mean and standard deviation Phone Error Rate (PER) on the TIMIT test set.
  } \label{tab:moments}
\end{minipage}
\end{table}

The Friedman test showed no significant differences between distributions of
PER across kaldifb, f-bank, g-bank, and tone-bank features ($Q=5.93,p=0.12$).
\Cref{tab:moments} shows that the f-bank condition lead to the lowest mean
PER. A Wilcoxon signed-rank test found no significant difference between
distributions of PER across f-bank and sif-bank features ($W=13,p=0.26$).
Therefore we cannot conclude that one feature representation lead to better
average PERs overall.

Switching from Kaldi's f-banks to any of our filter banks improved PER by
approximately 0.1\%, but otherwise the filters were quite similar in effect.
Furthermore, the size of the variance between runs is large enough to discount
any gains or losses.

The average PER across runs on the test set was 18.72\%. This is higher than
the percentage reported in \cite{Zhang2016Towards} 18.2\%, but the authors
of said paper did not report an average over trials. One trial with sif-banks
yielded a PER of 18.21\%.

Experimentation excluding dithering tended to have lower overall PERs
of around 18.5\%, with the lowest observed rate from sif-banks at
17.89\%. Dithering inserts random noise into the signal, which is intended to
prevent systems from over-fitting on the signal. It is therefore possible to
reduce the overall PER of the systems by removing dithering, but this is not
recommended.

One must be careful about what is concluded from these results. They are
consistent with the null hypothesis, which suggests that swapping features
in end-to-end phone recognition has little effect on PER, we conjecture that
this is indeed the case. Given the recent interest and success of learned
filter bank representations, notably that of
\cite{zeghidourLearningFilterbanksRaw2018}, one might extend this perspective
and claim that exploring fixed filter banks is outdated and irrelevant. The
difficulty comparing this work on fixed filter banks and previous work on
learned filter banks (e.g. \cite{zeghidourLearningFilterbanksRaw2018,Ghahremani2016Acoustic,Sainath2015Learning})
is that the latter class of paper tends to present a single error rate when
discussing results, whereas we have presented a proper statistical analysis
of the results over repeated trials. It is unclear whether those individual
numbers are representative of a trend or a lucky setting of parameters. For
example, as mentioned, we observed the PER of 17.89\% (twice, in fact) from
sif-banks. Had we only presented the best results, short-integration would
have beaten out the best result (learned or otherwise) from
\cite{zeghidourLearningFilterbanksRaw2018}. We are not concluding
here that sif-banks are better than learned filter banks%
\footnote{%
  We question, however, why learned filter banks tend to perform best when
  their weights are initialized with coefficients from a fixed filter bank, and
  why the learned banks tend to so strongly resemble a fixed filter bank in
  weight distribution, if what they learn is worth the additional effort and
  parameters.
} or even traditional f-banks. The statistical analysis does not support that,
nor does it seem likely. But neither can we come to understand more than what
is merely possible from individual error rates.

\section{Conclusions}
We presented two distinct ways of modifying traditional f-bank
features in order to produce a higher-resolution time-frequency
representation of a speech signal. The first is to replace triangular
filters with Gabor filters, which have a theoretically optimal
time-bandwidth product, or Gammatone filters, which are more faithful to
human speech perception. The second is to window over the filter
response only after taking its power. This leverages the low-pass
characteristic of the modulus to minimize information loss after
windowing. Experimentation with a modern end-to-end CNN architecture with CTC
\cite{Zhang2016Towards} yielded no significant effect of filters or the
modification to computation on PER. All filter bank and computation
implementations are available as an open-source Python package
as well as through the open-source Matlab repository COVAREP
\cite{g.degottexCOVAREPCollaborativeVoice2014}.

All of our results are predicated on swapping out STFT features for
short-integration features without any adjustments to the model architecture.
End-to-end neural ASR backpropagates through the decoding process. Even at the
level of phone recognition, end-to-end models
are responsible for modeling an entire sequence. Contrast this with an
acoustic model in a hybrid DNN-HMM, which is only responsible for
(and trained for) discriminatively classifying each frame of features. In the
latter case, decisions are a function of only the features within a small
context window, whereas an end-to-end system uses the entire utterance context.
It is likely that an end-to-end system might spend more resources building
a probability distribution over sequences of phones than it does focusing on
any interval of speech features, nullifying the differences between filter
types and resolution. In our experience, there is considerable variability in
the outcome of the training process of end-to-end systems. More experimentation
with traditional, hybrid architectures could prove illuminating in this
respect.

Another interesting question is whether a neural architecture exists that is
capable of
processing shorter frame intervals. As was mentioned in \Cref{sec:si}, frame
shift and window size can drastically affect the resolution in time of the
feature representation. Where the support in time of the square root filters
in a Mel-scaled f-bank range between about 90-400ms, Mel-scaled
complex Gabor filters with roughly the same bandwidths in frequency range
between about 3-24ms in support in time. Gabor filters are much more
capable of representing information at much smaller time scales. Unfortunately,
existing DNNs already suffer from vanishing and exploding gradients when
dealing with long time series data
\cite{pascanuDifficultyTrainingRecurrent2013}. The end-to-end CNN experimented
on here relies on delta and double-delta features to transmit long-term
dependency information. Increasing the total number of frames in an utterance
would make it more difficult to transmit this data. Also, averaging over
shorter time intervals means less spectral smoothing, which may prove difficult
for architectures that are not noise robust. Such a feature set may nonetheless
be a more user-friendly and robust alternative to backpropagating directly
through the speech signal, without requiring the additional parameters.

\section{Acknowledgements}
This research was funded by a Canada Graduate Scholarship and a Strategic
Project Grant from the Natural Sciences and Engineering Research Council of
Canada.

\printbibliography

\end{document}